\documentclass{article}

\usepackage{etoolbox}
\usepackage{cite}
\usepackage{arxiv}
\usepackage{amsmath}
\usepackage[utf8]{inputenc} % allow utf-8 input
\usepackage[T1]{fontenc}    % use 8-bit T1 fonts
\usepackage{hyperref}       % hyperlinks
\usepackage{url}            % simple URL typesetting
\usepackage{booktabs}       % professional-quality tables
\usepackage{amsfonts}       % blackboard math symbols
\usepackage{nicefrac}       % compact symbols for 1/2, etc.
\usepackage{microtype}      % microtypography
\usepackage{lipsum}
\usepackage[pdftex]{graphicx}
\usepackage{comment}
\usepackage{here}
\usepackage{svg}
\usepackage{overpic}
\usepackage[symbol]{footmisc}
\usepackage{caption}

\graphicspath{ {./images/} }

\makeatletter
\renewcommand{\@fnsymbol}[1] {%
  \ifcase#1\or
  1\or 2\or 3\or 4\or 5\or 6\or *
  \else\@ctrerr\fi}
\makeatother

{\title{\rm FinchGPT: A Transformer-based Language Model for Birdsong Analysis}}

\author{
	 Kosei Kobayashi\thanks{Graduate School of Life Sciences, Tohoku University,Katahira 2-1-1, Aoba-ku, Sendai, Miyagi 980-8577, Japan.}\\
	   \And
	 Kosuke Matsuzaki\thanks{Graduate School of Information Sciences, Tohoku University, 6-6-05 Aramaki Aza Aoba, Aoba-ku, Sendai, 980-8579, Japan.}\\
	   \And
	 Masaya Taniguchi\thanks{RIKEN Center for Advanced Intelligence Project, RIKEN, 1-4-1 Nihonbashi, Chuo-ku, Tokyo, 103-0027, Japan.}\\
	   \And
	 Keisuke Sakaguchi\footnotemark[2]\footnotemark[3]\\
	   \And
	 Kentaro Inui\thanks{Natural Language Processing Department, Mohamed bin Zayed University of Artificial Intelligence,}\thanks{Center for Language AI Research, Tohoku University, 6-6-05 Aramaki Aza Aoba, Aoba-ku, Sendai, 980-8579, Japan.}\footnotemark[3]\\
	   \And
	 Kentaro Abe\footnotemark[1]\footnotemark[5]\thanks{Division for the Establishment of Frontier Sciences, Tohoku University, Katahira 2-1-1, Aoba-ku, Sendai, Miyagi 980-8577, Japan}
\thanks{\textbf{Corresponding author} Kentaro Abe, Ph. D., Email address: k.abe@tohoku.ac.jp}
	}
	
\begin{document}

\maketitle

%アブスト
\begin{abstract}
The long-range dependencies among the tokens, which originate from hierarchical structures, are a defining hallmark of human language. However, whether similar dependencies exist within the sequential vocalization of non-human animals remains a topic of investigation. Transformer architectures, known for their ability to model long-range dependencies among tokens, provide a powerful tool for investigating this phenomenon. In this study, we employed the Transformer architecture to analyze the songs of Bengalese finch (\textit{Lonchura striata domestica}), which are characterized by their highly variable and complex syllable sequences. To this end, we developed FinchGPT, a Transformer-based model trained on a textualized corpus of birdsongs, which outperformed other architecture models in this domain. Attention weight analysis revealed that FinchGPT effectively captures long-range dependencies within syllables sequences. Furthermore, reverse engineering approaches demonstrated the impact of computational and biological manipulations on its performance: restricting FinchGPT's attention span and disrupting birdsong syntax through the ablation of specific brain nuclei markedly influenced the model's outputs. Our study highlights the transformative potential of large language models (LLMs) in deciphering the complexities of animal vocalizations, offering a novel framework for exploring the structural properties of non-human communication systems while shedding light on the computational distinctions between biological brains and artificial neural networks.
\end{abstract}

% keywords can be removed

%イントロ
\section{Introduction}
Recent studies in natural language processing utilizing deep neural networks have demonstrated striking increment on accuracy in predicting human speech patterns~\cite{1, 2, 3, 4, 5}. However, the applicability of these models to communication signals beyond human language remains largely unexplored. 
A commonly accepted characteristic of human speech is its variability, wherein tokens are arranged into sequence according to hierarchical rules that govern their order~\cite{6, 7, 8}. Given that the vocalization of many species is composed of short and limited vocal elements or syllables, animal vocalization has traditionally been assumed to lack such complex rules~\cite{6, 9}. For example, human language commonly exhibits long-range syntactic dependencies, such as those between a subject and its verb, whereas the vocalizations of non-human animals are generally believed to be constrained by simple transition rules among the sound elements, often describable by Markov processes~\cite{9, 10}. Nevertheless, some species employ a diverse array of syllables and variable sequences to communicate, suggesting the existence of dependencies among non-adjacent tokens~\cite{11}. 
Of particular interest are Bengalese finch (\textit{Lonchura striata domestica}), which utters 20–30 distinct syllables arranged in highly variable sequences, often referred to as songs~\cite{12}. These finches can be bred and maintained under laboratory conditions, making them ideal subjects for recording large volumes of audio data necessary for the sequential analysis of their songs. The syllables produced by Bengalese finches are clear and relatively easy to annotate into text data, especially with the aid of recently developed machine learning-based tools, such as the SAIBS program~\cite{13}. 
Theories of language understanding traditionally process syntactic and semantic information separately, employing distinct computational operations for each. Whether a semantic component exists within the sequence of syllables in birdsongs remains an unresolved and contentious question. In contrast, current Transformer-based architectures offer a unique advantage in this context, as they inherently integrate semantic and syntactic information without necessitating explicit separation. Building on this premise, we hypothesized that Transformer-based language models could serve as a powerful framework for decoding the sequential organization of syllables in birdsongs, potentially uncovering deeper insights into their structural complexity. 
In this study, we compiled a comprehensive song corpus of Bengalese finches and applied various language models and neural network architecture to generate and analyze their sequence of syllables used for communication.

\section{Related work}
With the advancement and widespread adoption of neural network models, some studies have explored their application to the vocalization of non-human animals. Many of such studies employed neural networks to effectively embed animal sounds into lower dimensional spaces, facilitating species detection from their vocal signals~\cite{14, 15, 16}. Some have utilized Transformer-based models, such as HuBERT~\cite{17} and wav2vec 2.0~\cite{18}, enabling more expressive encoding and classification~\cite{19, 20, 21, 22}. 
These previous efforts have predominantly concentrated on analyzing the phonological features of individual syllables. In contrast, this present study emphasizes the sequential relationship among syllables by applying language models to tokenized birdsongs. While transcribing audio sequence into text reduces the overall information contents, it enables analyses focused on sequential patterns, providing novel insights into the structural aspect of communication through the birdsong, a dimension previously overlooked in non-human vocal communication.

\section{Materials and Methods}
	\paragraph{Animals}
Bengalese finches (\textit{Lonchura striata domestica}) used in this study were purchased from Asada Chojyu Tradings and kept on a 14-h/10-h light cycle (daytime: 8:00--22:00). The care and experimental manipulation of animals used in this study were reviewed and approved by the institutional animal care and use committee of Tohoku University (2020LsA-005-07). All experiments and maintenance were performed following relevant guidelines and regulations.

	\paragraph{Song Recording and texturization}
Song recording and texturization was performed as described before~\cite{13}. Briefly, the subject birds were isolated in a cage in a soundproof chamber (68 cm × 51 cm × 40 cm), and their songs were recorded through a microphone (ECM8000, Behringer), digitalized with a sampling rate of 44.1 kHz, 16 bits by OCTACAPTURE (UA-1010, Roland). Songs are extracted by SAP2011~\cite{23} and stored in the song corpus of each bird with a timestamp. 
To texturize the recorded song, we utilized an automated syllable annotator SAIBS~\cite{13}. Training of SAIBS was performed for each bird using the 11,000 recorded song files for each bird. For syllable segmentation, we set the fast Fourier transform (FFT) window as 512 samples at 44.1 kHz. To determine whether the sequence of the extracted syllables is from songs and not from movement-related sounds or noise, we introduced a cutoff value that increases when syllables are detected while regressing according to the time-lapse of the non-syllable segment. The end of songs was defined as 2 s after the detection of the final syllable. 
We collected the songs of 3 male Bengalese finches and created texturized song corpus for each. The length of the recording periods was 35, 105, and 30 days for each subject, which corresponded to 13,288, 4,492, and 14,435 songs, respectively. By converting those song sounds into text by SAIBS, song corpus of 1,173 kB, 236 kB, and 731 kB file size were created, respectively. Based on the holdout method, these corpora were split and used for training, evaluation, and testing in each subject.

	\paragraph{Procedures for training Markov, RNN, LSTM, and Transformer models}
Model training, evaluation, and testing were conducted on corpora created individually from each bird. The RNN, LSTM, and Transformer models were trained to minimize cross-entropy loss for next token prediction. The optimization algorithm used was Adam for RNN and LSTM models, and AdamW for the Transformer models, each with a learning rate of 0.001. The training process utilized a batch size of 32, and the training loops continued until the losses converged. All training was performed on a single GPU (Tesla T4 15GB, CUDA Version: 12.2) using Google Colaboratory. 
RNN, LSTM given optimal hyper parameters obtained by Tree-structured Parzen Estimator (TPE). 100 trials were performed by TPE and the combination with the lowest cross-entropy loss was adopted. Type and range of parameters optimized were: number of layers, 1--6; hidden dimension, 10--100; embedding dimension, 10--100; and dropout probability, 0.0--1.0. To prevent over-fitting, training was terminated if no performance improvement on evaluation data was observed over five consecutive epochs. 
For the Transformer-based language model, we employed an architecture equivalent to GPT-2~\cite{2}, varying the number of layers, hidden states, attenuation heads, and vocabulary sizes. In Fig \ref{fig:fig1}, the model consists of 6 layers, 384 hidden states, and 6 attention heads. The vocabulary size varies from 15 to 17 (including 4 special tokens), is specific to each individual. Only models trained exclusively on the finch corpus from a completely uninitiated state, with no prior learning or exposure to information about human language, were used in this study. 
The artificial song corpus obeying the 6th-order Markov rule was created as follows: First, a start token and an end token were defined and combined at the beginning and end of each natural song’s syllable sequence. Next, transition probabilities between tokens, including these special tokens, were calculated. To generate an artificial song, a start token was entered into the model, which then iteratively outputs the next token based on the transition probabilities until the end token was generated, completing a single song. During the initial stage of generation, when the context size was less than six, the generation was defaulted to a Markov model corresponding to the largest available context size.

	\paragraph{Attention head analysis and Attention span restriction}
Attention head analysis was conducted on FinchGPT-middle with configuration of 6L, 6A, 384H. The sample song used in analysis shown in Fig \ref{fig:fig2}A is the same as the song shown in Fig \ref{fig:fig2}B, with special tokens appended at the beginning and end. Dependency parse tree were derived from the attention weights in layer 1 and 6 using the Chu-Liu-Edmonds algorithm to extract maximum spanning trees~\cite{24, 25}. The average attention span length was calculated by extracting token pairs with attention weights greater than 0.5 in each self-attention head and computing the mean of their distances across 200 songs. For attention span restrictions, masking was applied to unreferenced tokens through all layers. Both training and testing were performed under these restricted attention conditions. 

	\paragraph{Song classification task}
Finetuning was performed on FinchGPT that was pre-trained on the before HVC-ablation corpus, for a classification task. The training dataset consisted of songs labeled as either before- or after HVC-ablation, with a size of 139 kB, corresponding to 1,200 before-ablation songs and 880 after-ablation songs. The evaluation dataset comprised 100 songs from each category, while the test dataset included 200 songs each from before- and after- ablation categories. The upper displayed in Fig \ref{fig:fig4}F represents the accuracy achieved when testing on the training dataset. This upper bound accounts for the presence of ambiguous songs in the labeled corpus, which cannot be classified even under conditions of overfitting, thereby reflecting the proportion of such ambiguous song.

	\paragraph{Surgical procedures}
Adult Bengalese finch was anesthetized with a medetomidine-midazolam-butorphanol mixture (medetomidine 30 \textmu{}g/mL, midazolam 30 \textmu{}g/mL, butorphanol tartrate 500 \textmu{}g/mL, NaCl 118 mM; 200 \textmu{}L per bird), after which ibotenic acid (0.5 \textmu{}L, 2.0 mg/mL in PBS) was injected unilaterally for HVC at the coordinate from the Y sinus: Anterior, 0 mm; lateral, 2.2 mm; depth 0.75 mm at beak angle 40°.

	\paragraph{Model Evaluation Indicators}
We used accuracy and cross-entropy for the next token prediction task, and accuracy and AUC for the song-classification task as the evaluation metrics of the model. Accuracy for the next-token prediction task refers to the percentage of correct predictions made by the model. Specifically, it is calculated as the number of correct predictions divided by the total number of predictions made. Cross-entropy is calculated during the prediction of each token, except for the end token and the token immediately following the start token (i.e., the first syllable in a song). The average cross-entropy is computed without applying any weighting based on the frequency of token occurrences. 
The cross-entropy value for a token prediction is calculated as follows, given the true probability distribution $p$ and the probability distribution output by the language model $q$:

\[
      H(p, q) = - \sum_{x} p(x)\log_{}q(x)
\]

Accuracy for the song-classification task represents the percentage of correct binary classifications. This is calculated as follows using values of true positive (TP), false positive (FP), true negative (TN), and false negative (FN):

\[
\textit{Accuracy} = \frac{\text{TP} + \text{TN}}{\text{TP} + \text{FP} + \text{TN} + \text{FN}}
\]

The true positive rate and false positive rate for receiver operating characteristics analysis are calculated as follows:

\[
      \textit{True positive rate} = \frac{\text{TP}}{\text{TP} + \text{FN}} \qquad  \textit{False positive rate} = \frac{\text{FP}}{\text{FP} + \text{TN}}
\]

In Fig \ref{fig:fig4}F, the post HVC-ablation corpus was considered the positive class for these calculations.

	\paragraph{Acoustic feature analysis of HVC ablated bird}
For the acoustic analysis, all the syllables used by the subject bird in its songs were analyzed. For each syllable, 300 samples were randomly selected from the corpus before and after the HVC ablation. Pitch was calculated as the mean value of the peak frequency, which was defined as the frequency with the highest power at each time point throughout the duration of each syllable.
                
	\paragraph{Statistics}
Paired \textit{t}-test, Wilcoxon rank sum test and Wilcoxon signed-rank test were used for the comparison of two data sets. 
The criterion of statistical significance was set as \textit{P} < 0.05.
All statistical analyses were conducted using R (version 4.3.1).

%ﾘｻﾞﾙﾄ
\section{Results}
\subsection{Evaluation of LLM trained with birdsongs}
\begin{figure}
		  \begin{minipage}[b]{1\hsize}
		    \centering
     		           \raisebox{5mm}{
			   \begin{overpic}[width=0.06\textwidth]{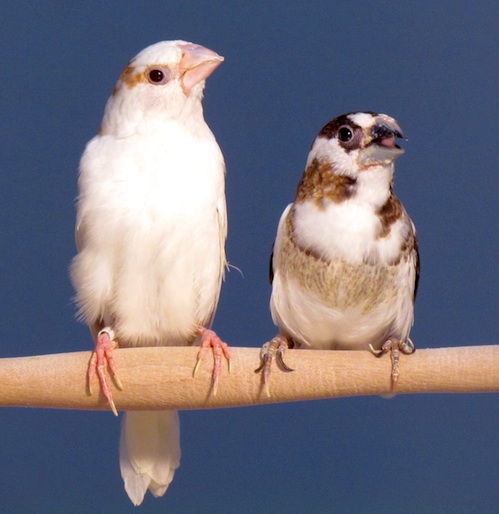}
			        \put(-12,170){\fontsize{10}{22}\selectfont\textbf{\textsf{A}}}
		   	   \end{overpic}
		   	   }
				\hspace{-2mm} % 横方向の空白を追加
     		           \raisebox{0mm}{
			   \begin{overpic}[width=0.43\textwidth]{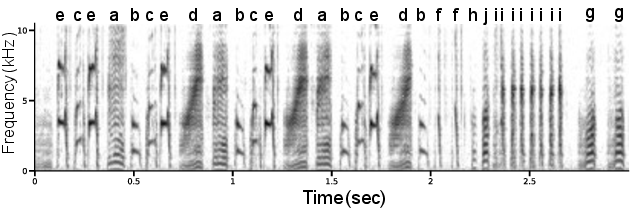}
			        \put(0,26){\fontsize{10}{22}\selectfont\textbf{\textsf{}}}
		   	   \end{overpic}
		   	   }
     			          \hspace{3mm} % 空白を追加
     			   \raisebox{2mm}{
			   \begin{overpic}[width=0.45\textwidth]{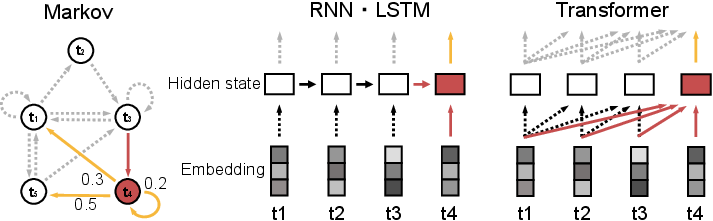}
			        \put(-3,28){\fontsize{10}{22}\selectfont\textbf{\textsf{B}}}
		   	   \end{overpic}
		   	   }
 		  \end{minipage}
 		  
 		          \vspace{5mm} % 図の間にスペースを追加
 		  
		  \begin{minipage}[b]{1\hsize}
		    \centering
				\hspace{0mm} % 横方向の空白を追加	  
		    	   \raisebox{0mm}{
			   \begin{overpic}[width=0.35\textwidth]{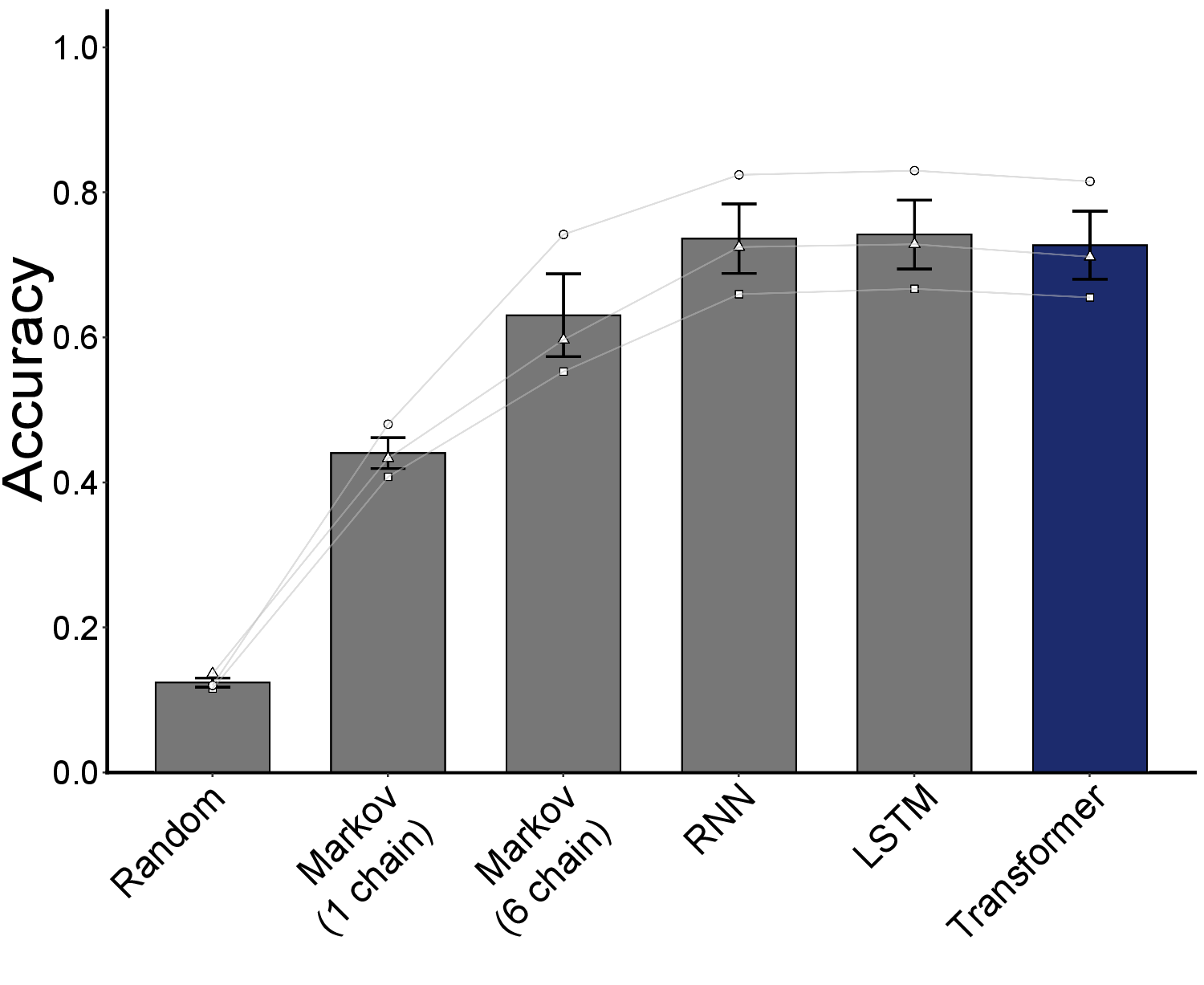}
			        \put(-4,80){\fontsize{10}{22}\selectfont\textbf{\textsf{C}}}
		   	   \end{overpic}
		   	   }
				\hspace{0mm} % 横方向の空白を追加
			    \raisebox{-1.8mm}{
			   \begin{overpic}[width=0.2\textwidth]{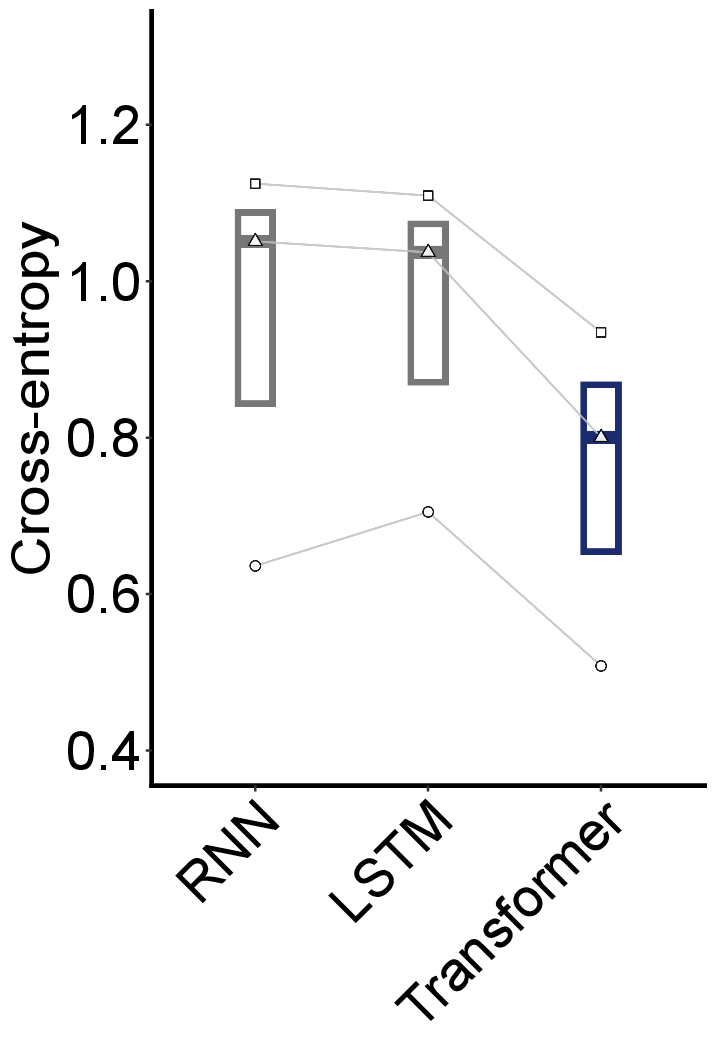}
			        \put(0,99){\fontsize{10}{22}\selectfont\textbf{\textsf{D}}}
		   	   \end{overpic}
		   	   }
				\hspace{0mm} % 横方向の空白を追加
			  \raisebox{-8.9mm}{
			   \begin{overpic}[width=0.39\textwidth]{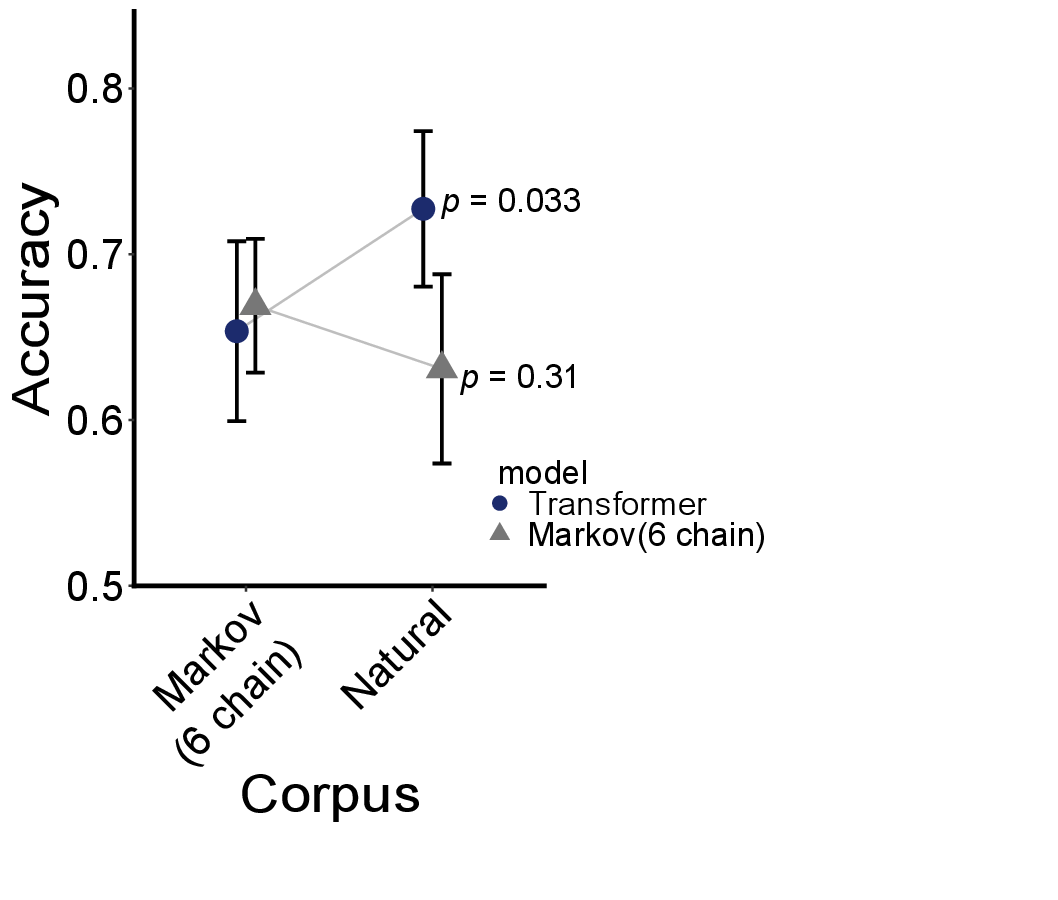}
			        \put(0,85.5){\fontsize{10}{22}\selectfont\textbf{\textsf{E}}}
		   	   \end{overpic}
		   	   }
 		  \end{minipage}
			
			\vspace{-3mm} % 図の間にスペースを追加
			
	  	\begin{minipage}[c]{1\hsize} 
  		    \centering
  		            \raisebox{0mm}{
			   \begin{overpic}[width=0.13\textwidth]{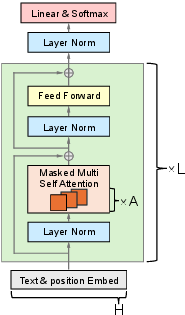}
			        \put(-14,95){\fontsize{10}{22}\selectfont\textbf{\textsf{F}}}
		   	   \end{overpic}
		   	   }
			   \hspace{3mm} % 横方向の空白を追加
		   	   \raisebox{2mm}{
			   \begin{overpic}[width=0.36\textwidth]{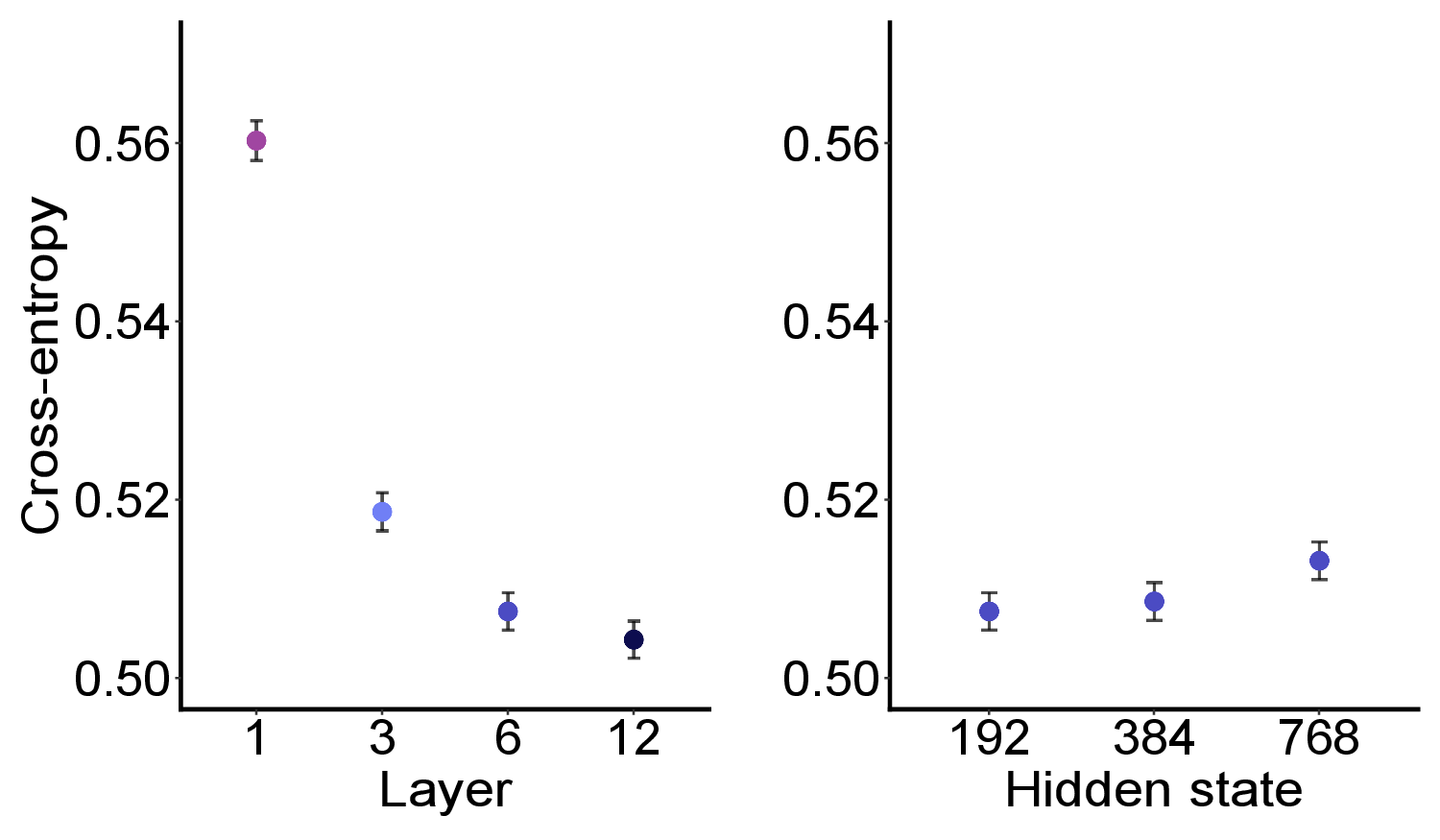}
			        \put(0,55){\fontsize{10}{22}\selectfont\textbf{\textsf{G}}}
		   	   \end{overpic}
		   	   }
		   	   \raisebox{-0.4mm}{
			   \begin{overpic}[width=0.4\textwidth]{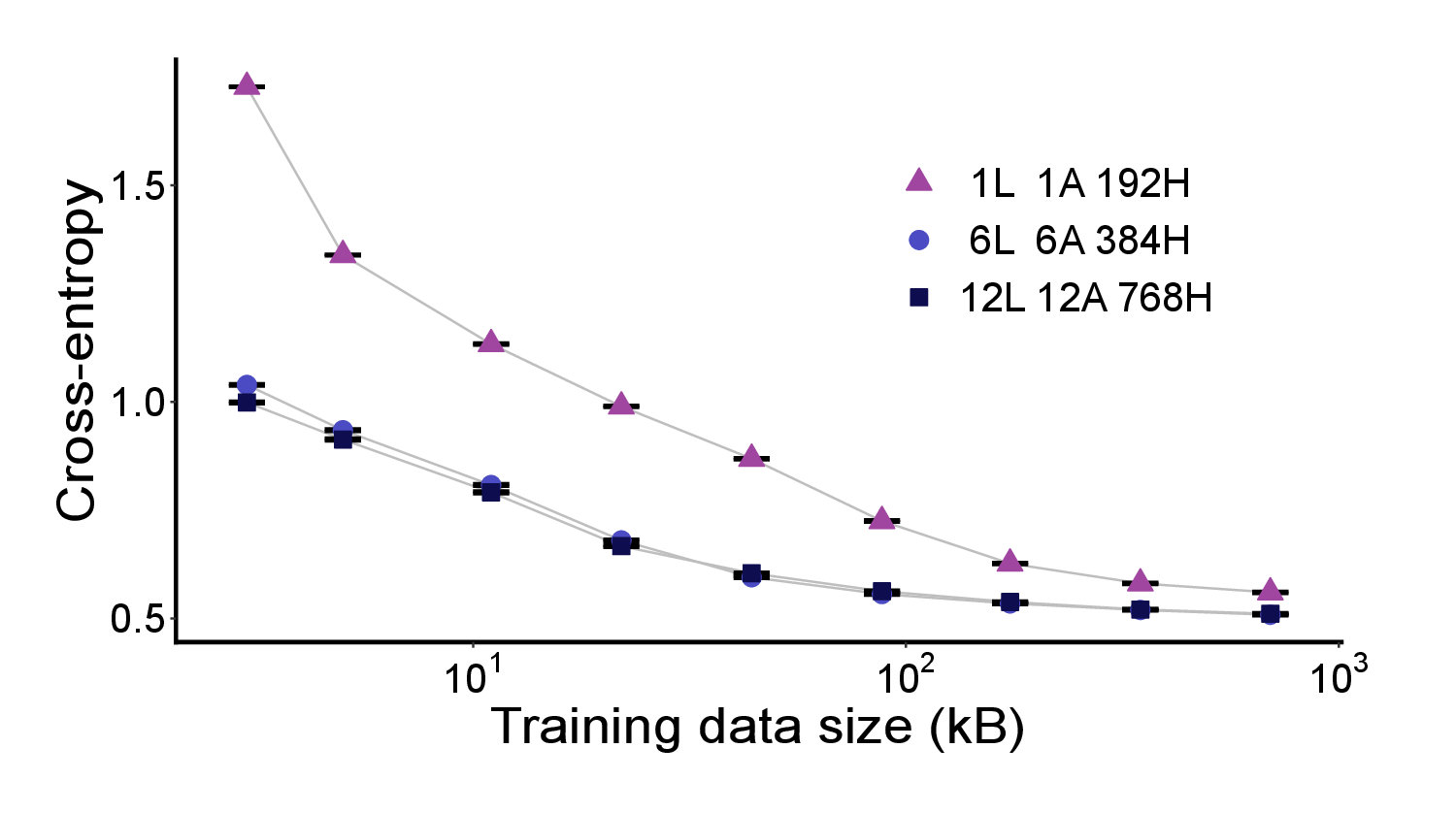}
			        \put(1,53){\fontsize{10}{22}\selectfont\textbf{\textsf{H}}}
		   	   \end{overpic}
		   	   }
	  	\end{minipage}
		    \caption{
\textbf{Generation and evaluation of GPT trained with birdsong.}
\textbf{(A)} A picture of Bengalese finch and a spectrogram of an example song of Bengalese finch, each syllable labelled with arbitrary alphabet symbols.
\textbf{(B)} Schematic diagrams showing the core computation concept of each algorithm in token processing. t1–t5 represent sequence of tokens input. Red highlights the token currently being processed and the information utilized for computation. The Markov model only predicts the next token according to the transition probabilities. RNN and LSTM propagate information primarily through hidden states from nearby tokens, while Transformer captures global context via self-attention. 
\textbf{(C)} Percentage of correct answers in the next token prediction task (mean ± SEM, n = 3 corpus each from different bird).
\textbf{(D)} Cross-entropy between model outputs and correct tokens. Box plot shows median and first and third quantiles. Actual data points representing each corpus are shown in gray. 
\textbf{(E)} Performance changes with artificially structured training and test corpora (mean ± SEM, n = 3 corpus each from different bird). P values, Paired t-test, n = 3.
\textbf{(F)} Diagram of the FinchGPT architecture. L, A, and H represent the number of layers, attention heads, and hidden state dimensions, respectively. The embedded dimensions size is identical to the hidden state. 
\textbf{(G)} Comparison of cross-entropy in the next token prediction task performed on FinchGPT with varying model parameters: layer (left, A = L, H = 192), and hidden state dimensions (right, L = 6 and A = 6).
\textbf{(H)} Same as (G) with varying training text data sizes. FinchGPT-small (1L, 1A, 192H), medium (6L, 6A, 384H), and large (12L, 12A, 768H) were shown. Mean ± SEM, n = 235,256 predictions across 2,660 songs for both comparisons for (G–H).
}
		    \label{fig:fig1}
		 \end{figure}

The text corpus was created by recording the vocalization of Bengalese finches and texturizing them using a machine leaning based bird-syllable annotator, SAIBS (Fig \ref{fig:fig1}A)~\cite{13}. For this study, we trained the algorithm with a text corpus derived from a single Bengalese finch, with each syllable treated as a token. Four models are evaluated: Markov, RNN, LSTM, and Transformer-based language model (Fig \ref{fig:fig1}B). To assess their performance, we executed a next-token (syllable) prediction task using the bird vocalization data. 
Figure \ref{fig:fig1}C compares accuracy for next-syllable prediction across the algorithms: Markov, RNN, LSTM, and Transformer. Cross-entropy analysis for next-syllable prediction reveals that the Transformer model achieved the lowest entropy score, indicating greater predictive performances compared to other models such as RNN and LSTM tested in this study (Fig \ref{fig:fig1}D). To explore this further, we extended the Markov model to account for up to six preceding syllables, yet Transformer consistently outperformed it (Fig \ref{fig:fig1}C). 
Next, using syllables from the same bird, we generated an artificial song corpus adhering to the 6th-order Markov rule and compared the performance of both the Markov and Transformer algorithms. The song corpus, produced by a 6th-order Markov model, inherently lacked dependencies between syllables beyond the immediately preceding six syllables. When evaluated on this corpus, the accuracy of the Transformer model declined significantly, whereas the Markov algorithm maintained or increased its performance (Fig \ref{fig:fig1}E). 
Long-range dependencies in the syllable sequences of birdsongs have been previously hypothesized, but their precise nature remains unclear~\cite{26, 11, 10, 27, 28}. The observed superior performance of the Transformer model, compared to Markov and other neural network models, suggests that long-range rules play a significant role in structuring birdsong, with attention-based architectures excelling at modeling complex sequential dependencies by efficiently capturing and integrating such rules, similar to those observed in human language~\cite{1, 3}. Hereafter, we refer to the Transformer trained on the birdsong corpus as FinchGPT (Finch version of Generative Pre-trained Transformer) and further analyzed their properties.

\captionsetup[figure]{labelfont=bf}

%		\begin{figure}[H]

\subsection{Evaluation of model parameters and corpus size}
The performance of transformer-based language models on human language is influenced by the number of model parameters, the size of the training data, and the computational resources allocated, a relationship often referred to as the scaling law~\cite{29}. However, the optimal configuration of parameters for modeling bird vocalization remains unknown. Thus, we investigated the effect of model size and training data size on the performance of FinchGPT. 
First, with the 695 kB text corpus, we analyzed the model performance across different layer sizes (L), attention heads (A), and hidden state dimensions (H) (Fig \ref{fig:fig1}F). The model engineered one attention head layer (1L, 1A) resulted in significantly worse cross-entropy (Fig \ref{fig:fig1}G). We observed a steady reduction of cross-entropy by increasing the layer numbers. In contrast, increasing the number of hidden states, ranging from 192 to 768, did not significantly affect cross-entropy (Fig \ref{fig:fig1}G). Based on these results, we established several FinchGPT models with varying parameters: FinchGPT-small (1L, 1A, 192H), medium (6L, 6A, 384H) and large (12L, 12A, 768H), which we used for subsequent analyses. 
Next, we trained FinchGPTs models (small, medium, large) on datasets that varied by approximately a factor of two in size. As data size increased, cross-entropy steadily decreased in all models (Fig \ref{fig:fig1}H). The medium and large models consistently achieved lower cross-entropy compared to the small-model across all dataset sizes. In the medium and large models, the rate of loss reduction when doubling the data size peaked around 30 kB, a size obtained by approximately 40 days of song recordings of single bird. While increased data collection enhances model accuracy, our observation indicates that this dataset size is sufficient for analyzing the birdsongs of most individuals.

\subsection{Attention head analysis}
\begin{figure}
		    \centering

			   \begin{overpic}[width=0.7\textwidth]{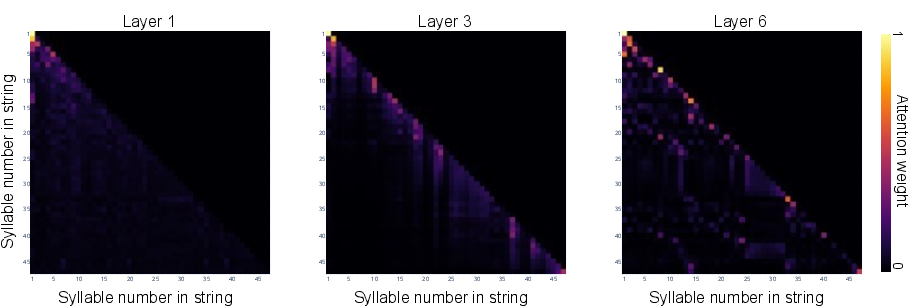}
			        \put(1,32){\fontsize{10}{22}\selectfont\textbf{\textsf{A}}}
		   	   \end{overpic}
			   \begin{overpic}[width=0.6\textwidth]{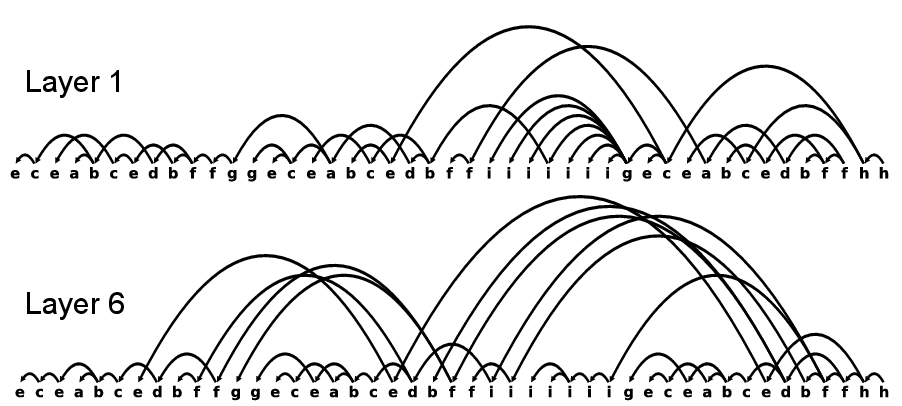}
			        \put(-7,39){\fontsize{10}{22}\selectfont\textbf{\textsf{B}}}
		   	   \end{overpic}
			\vspace{3mm} % 図の間にスペースを追加

		  \begin{minipage}[c]{1\hsize}
		    \centering
			   \begin{overpic}[width=0.23\textwidth]{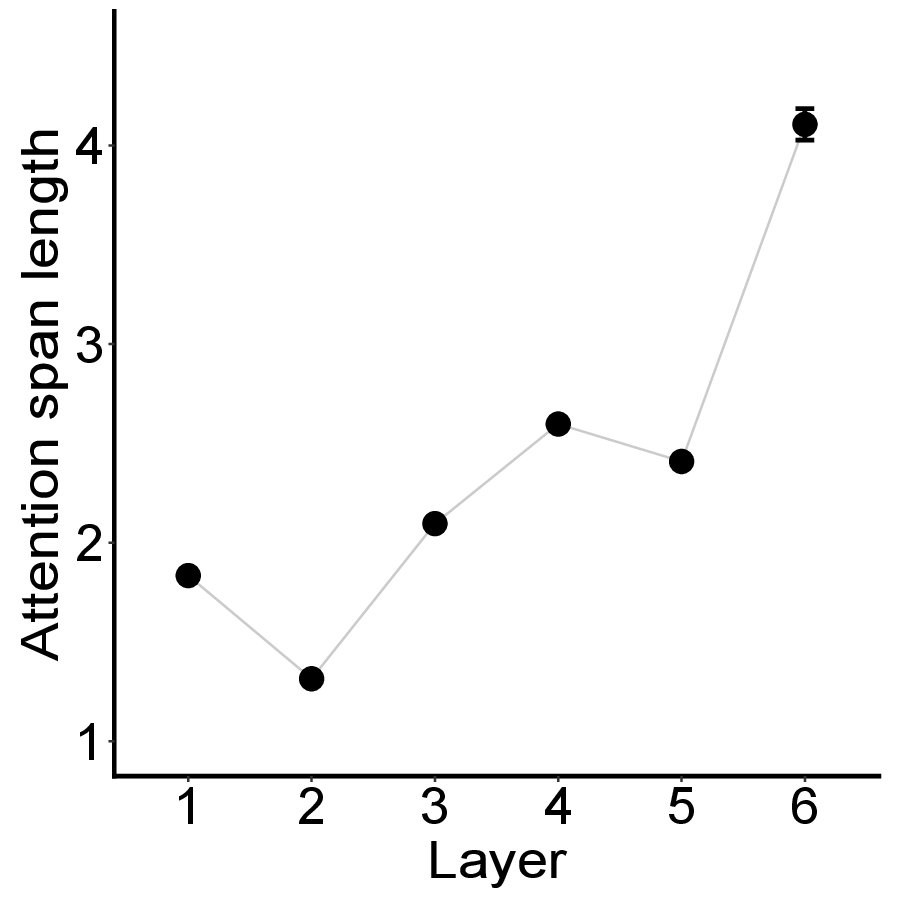}
			        \put(-3,95){\fontsize{10}{22}\selectfont\textbf{\textsf{C}}}
		   	   \end{overpic}	
		   	   \hspace{1mm} % 図の間にスペースを追加	
		   	   \raisebox{-0.5mm}{	   
			   \begin{overpic}[width=0.42\textwidth]{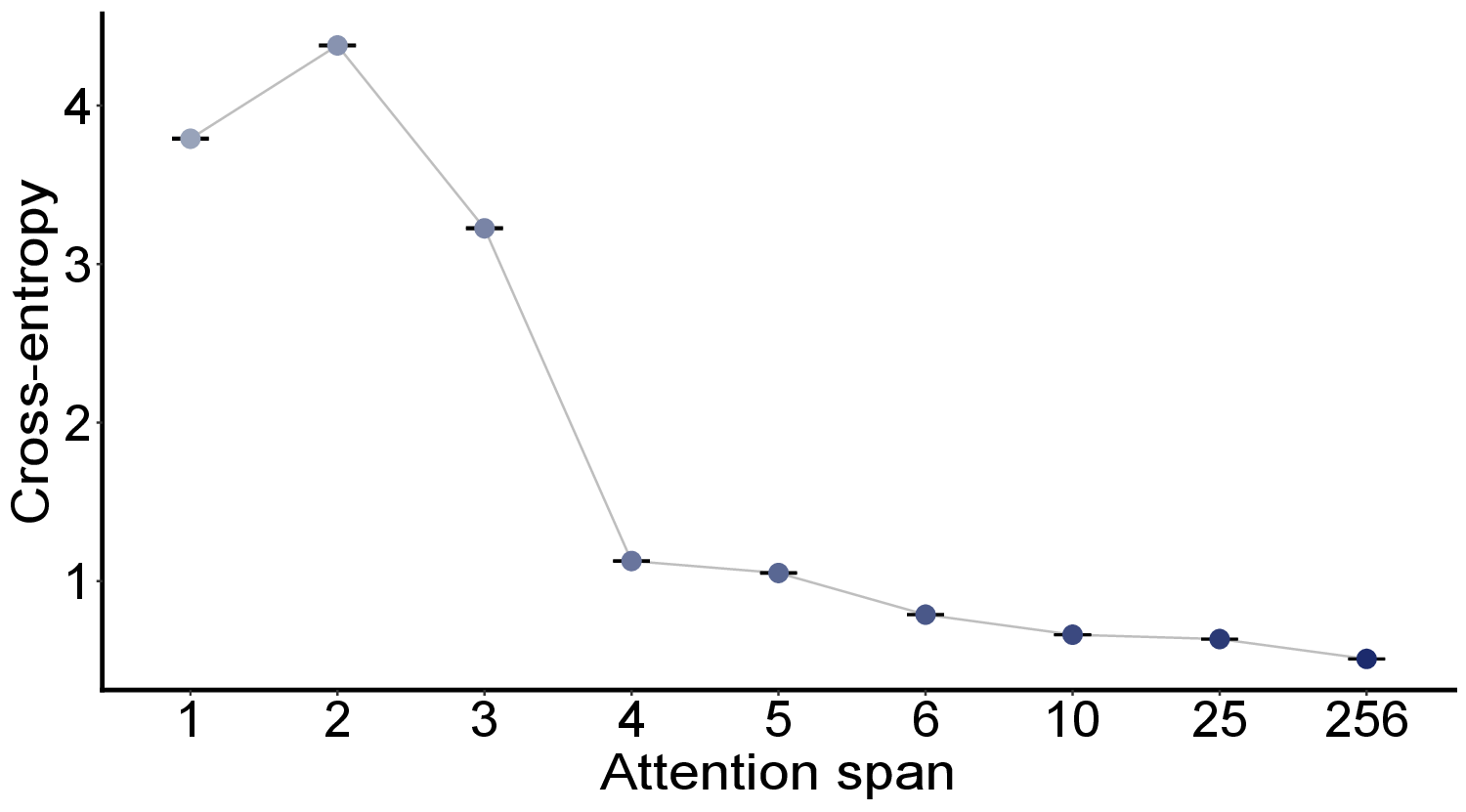}
			        \put(-1,53){\fontsize{10}{22}\selectfont\textbf{\textsf{D}}}
		   	   \end{overpic}
		   	   }
		  \end{minipage}

		    \caption{
\textbf{Attention visualization reveals the long-range dependencies within a song.}
\textbf{(A)} An Example of self-attention matrices within FinchGPT-medium (6L, 6A, 384H) when processing a song with 46 tokens (44 syllables + 2 special tokens) as input. The horizontal and vertical axes represent the order of input syllables, while the attention wights between syllables are color coded to indicate their relative intensities. 
\textbf{(B)} Example of observed attention in layer 1 and 6. Arrows represent the attention weights forming the maximum spanning tree, as identified by Chu-Liu-Edmonds algorithm. 
\textbf{(C)} Average span length of token pairs in each layer. Attention weights greater than 0.5 are analyzed (mean ± SEM, n = 2,935, 1,403, 4,124, 17,054, 22,670, and 11,080 from layer 1 to 6).
\textbf{(D)} The cross-entropy values of the model restricted to specific attention span lengths (mean ± SEM, n = 235,256 prediction across 2,660 songs). The numbers on x-axis indicate the maximum number of immediately preceding tokens that can be attended to in each attention operation. Unrestricted FinchGPT utilizes an attention span length of up to 256 tokens.
}
		    \label{fig:fig2}
		 \end{figure}

A key advantage of attention-based architecture is their capacity to visualize influential tokens during processing, offering enhanced interpretability compared to previous models including RNN-base models, whose internal operations are often opaque. Through attention visualization and intermediate layer probing, it has been demonstrated that in transformer models of human language, shallow layers primarily captures superficial relationships or short-range dependencies, while deeper layers encode more abstract, longer-range syntactic structures between tokens~\cite{30, 31}. 
By visualizing the attention heads of FinchGPT, we identified the certain heads focused on non-adjacent syllables to predict the next coming token (Fig \ref{fig:fig2}A, \ref{fig:fig2}B). Layer-wise analysis revealed that the average span-length of attention gradually increases as the layer deepens (Fig \ref{fig:fig2}C), suggesting that the ability to capture longer-range dependencies in the deeper layers is critical for accurate sequence prediction.
Notably, attention visualization using the Chu-Liu-Edmonds method~\cite{24, 25} revealed that in deeper layers, some attention spans extend beyond motifs, chunks of syllables that frequently appear together in conjunction (Fig \ref{fig:fig2}B)~\cite{32}. 
To assess the functional relevance of this long-range attention, we constrained the model’s attention span by limiting the range of referenceable tokens. The model with restricted attention span exhibited a marked decline in performance compared to those with unrestricted attention (Fig \ref{fig:fig2}D), underscoring the critical role of long-range dependencies in maintaining prediction accuracy of the birdsongs.

\subsection{Token embedding visualization}
\begin{figure}
		    \centering
			   \begin{overpic}[width=0.90\textwidth]{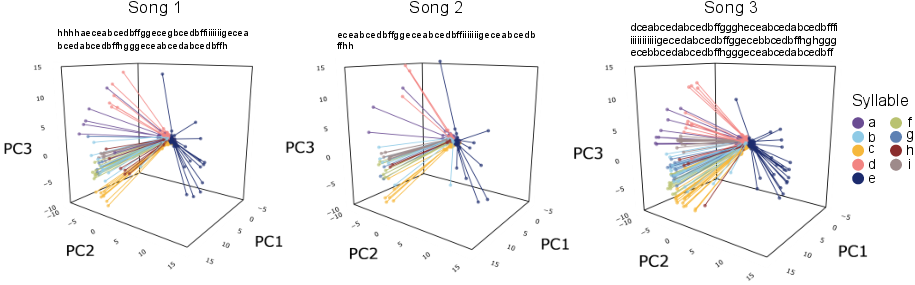}
			        \put(-1,30){\fontsize{10}{22}\selectfont\textbf{\textsf{A}}}
		   	   \end{overpic}	
		   	   
				\vspace{2mm} % 横方向の空白を追加
		    \centering
			   \raisebox{-4mm}{
			   \begin{overpic}[width=0.25\textwidth]{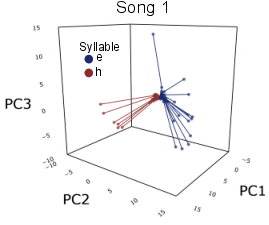}
			        \put(0,80){\fontsize{10}{22}\selectfont\textbf{\textsf{B}}}
		   	   \end{overpic}	
		   	   }			   
				\hspace{0mm} % 横方向の空白を追加
		   	   \raisebox{-2mm}{
			   \begin{overpic}[width=0.64\textwidth]{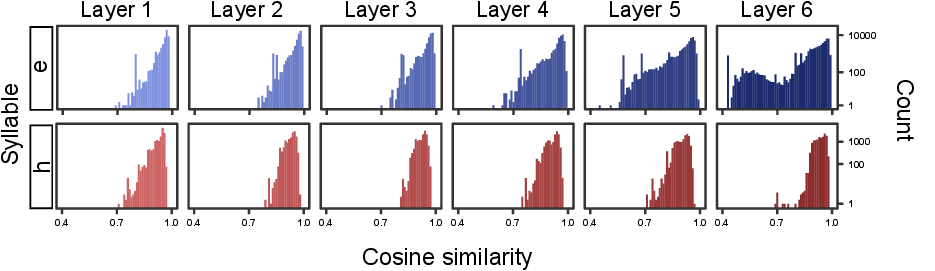}
			        \put(0,29){\fontsize{10}{22}\selectfont\textbf{\textsf{C}}}
		   	   \end{overpic}	
		   	   }
		    \caption{
\textbf{Token embedding analysis reveals the contextual usage of syllables.}
\textbf{(A)} Three-dimensional plots showing the trajectory of changes in token embeddings for three example songs across Transformer layers. The original 384-dimensional embeddings were reduced to 3 dimensions by PCA. Different colors represent distinct syllables. The three sample songs were randomly selected from the corpus of a single individual.
\textbf{(B)} Same as (A), only the tokens for syllable “e” and “h” are shown. 
\textbf{(C)} Distribution of cosine similarity relative to the mean embeddings of each layer in 384-dimensional space. The vertical axis is presented on a logarithmic scale.}
		    \label{fig:fig3}
		 \end{figure}

Attention-based models can capture context-dependent usage of tokens by progressively updating token embeddings across layers. In processing human language, this feature allows for the differentiation of homonyms---words with identical orthography but distinct meanings---with embeddings converging in different directions as they are refined layer by layer~\cite{33, 34, 35}. 
We visualized how syllable embeddings evolve across FinchGPT layers and analyzed their distribution properties (Fig \ref{fig:fig3}A). Using PCA to reduce the embedding dimensionality from 384 to 3, we found certain syllables, despite sharing identical phonological properties, followed by divergent embedding trajectories (Fig \ref{fig:fig3}B). 
We examined their cosine similarity distributions relative to their mean embeddings of each layer in 384 dimensions. We observed syllable dependent trajectories distributions. Specifically, for syllable “h,”  the cosine similarity distribution showed uniform pattern across the layers, but for the syllable “e,” the distribution become to exhibit a bimodal pattern as the layer deepens, suggesting the presence of two distinct embedding clusters corresponding to different contextual uses of the same syllable (Fig \ref{fig:fig3}C). Together, these analysis shows that FinchGPT captures the context dependent usage of syllable in birdsong. 

\subsection{Validation of performance by reverse engineering approaches}
\begin{figure}
		    \centering
		    	   \raisebox{5mm}{
			   \begin{overpic}[width=0.45\textwidth]{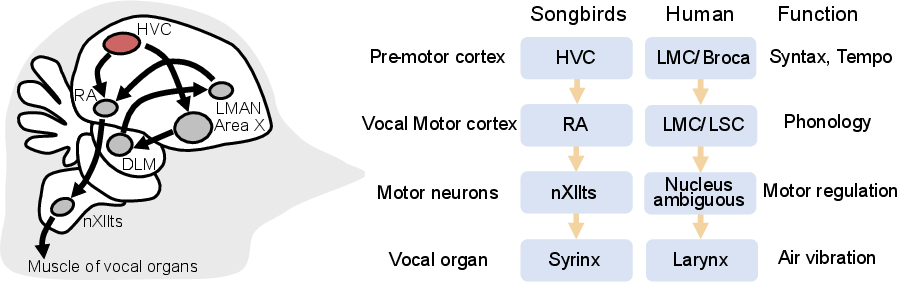}
			        \put(0,36.5){\fontsize{10}{22}\selectfont\textbf{\textsf{A}}}
%			        \vspace{5mm} % 横方向の空白を追加
		   	   \end{overpic}
		   	   }
		   	   \hspace{2mm} % 横方向の空白を追加
		   	   \raisebox{0mm}{
			   \begin{overpic}[width=0.3\textwidth]{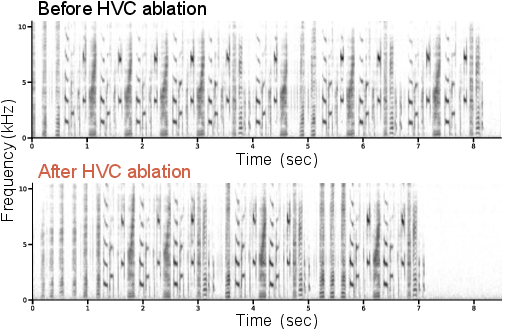}
			        \put(-7,65){\fontsize{10}{22}\selectfont\textbf{\textsf{B}}}
		   	   \end{overpic}	
		   	   }
		   	  
			\vspace{1mm} % 図の間にスペースを追加
						
	  	\begin{minipage}[c]{1\hsize} 
  		    \centering
  		    \hspace{0mm} % 横方向の空白を追加	
		   	   \raisebox{5.9mm}{	
			   \begin{overpic}[width=0.24\textwidth]{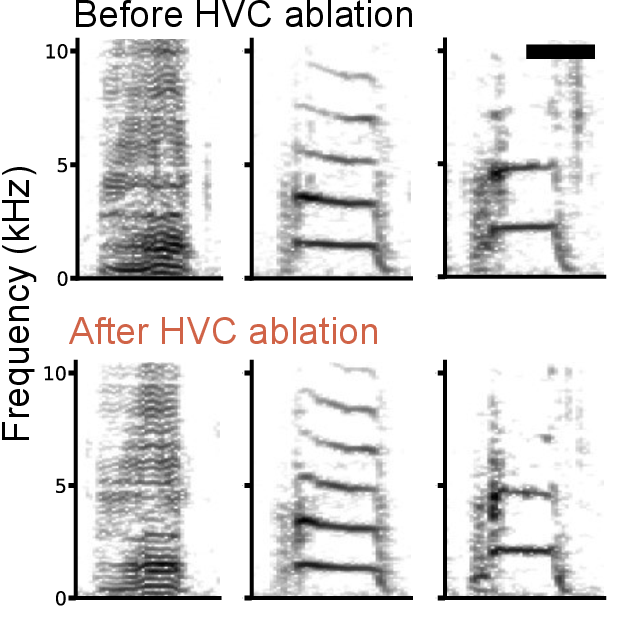}
			        \put(-5,96){\fontsize{10}{22}\selectfont\textbf{\textsf{C}}}
		   	   \end{overpic}	   	   	
		   	   }
			   \hspace{-3mm} % 横方向の空白を追加		   	   
		   	   \raisebox{2.0mm}{	
			   \begin{overpic}[width=0.17\textwidth]{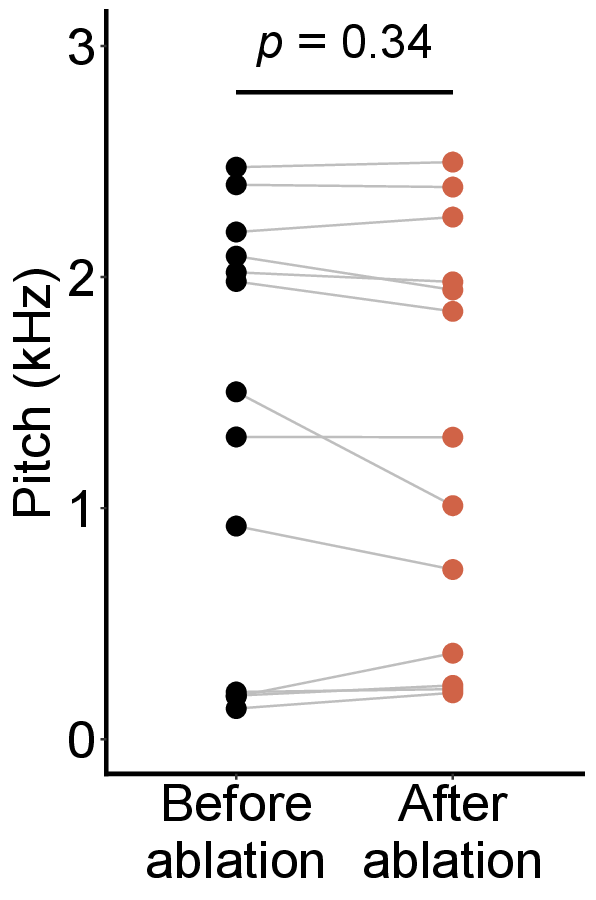}
			        \put(-5,50){\fontsize{10}{22}\selectfont\textbf{\textsf{}}}
		   	   \end{overpic}	   	   	
		   	   }
			    \hspace{2mm} % 横方向の空白を追加	
			    \raisebox{0mm}{
			   \begin{overpic}[width=0.22\textwidth]{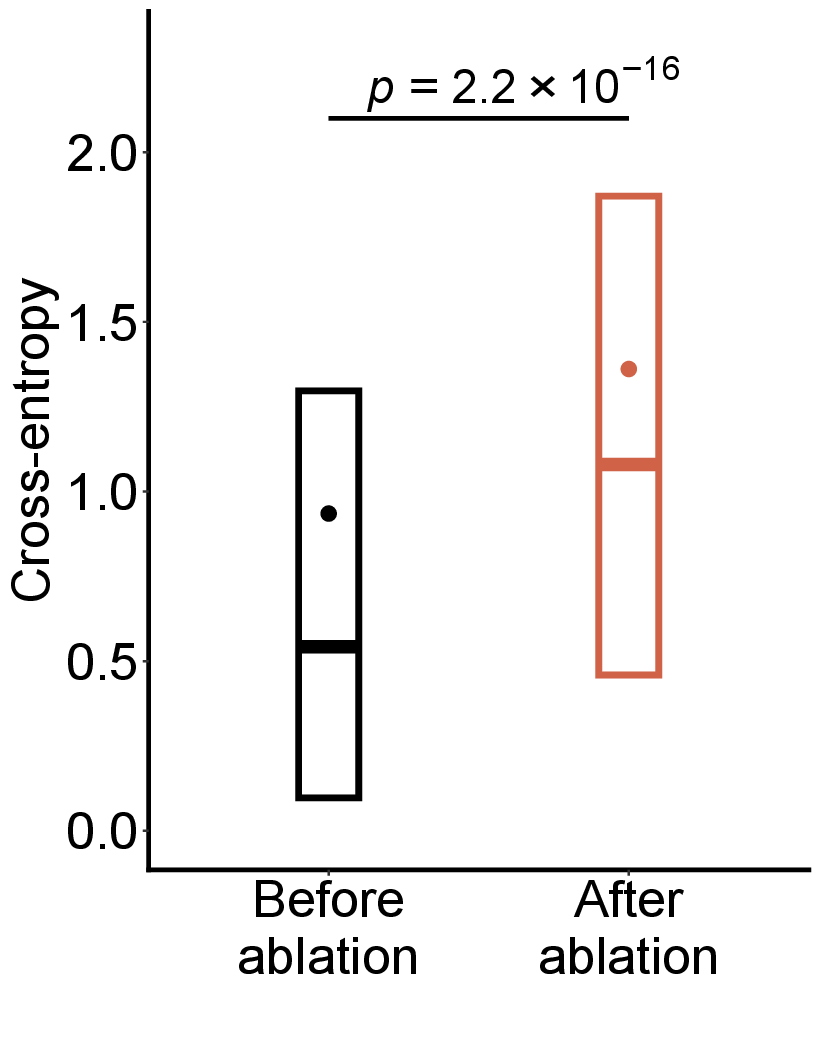}
			        \put(1,95){\fontsize{10}{22}\selectfont\textbf{\textsf{D}}}
		   	   \end{overpic}	
		   	   }
		   	   \hspace{15mm} % 横方向の空白を追加	
	  	\end{minipage}
	  	
		\vspace{1mm} % 横方向の空白を追加
	  	
	  	\begin{minipage}[c]{1\hsize} 
  		    \centering
				\hspace{0mm} % 横方向の空白を追加
			   \begin{overpic}[width=0.27\textwidth]{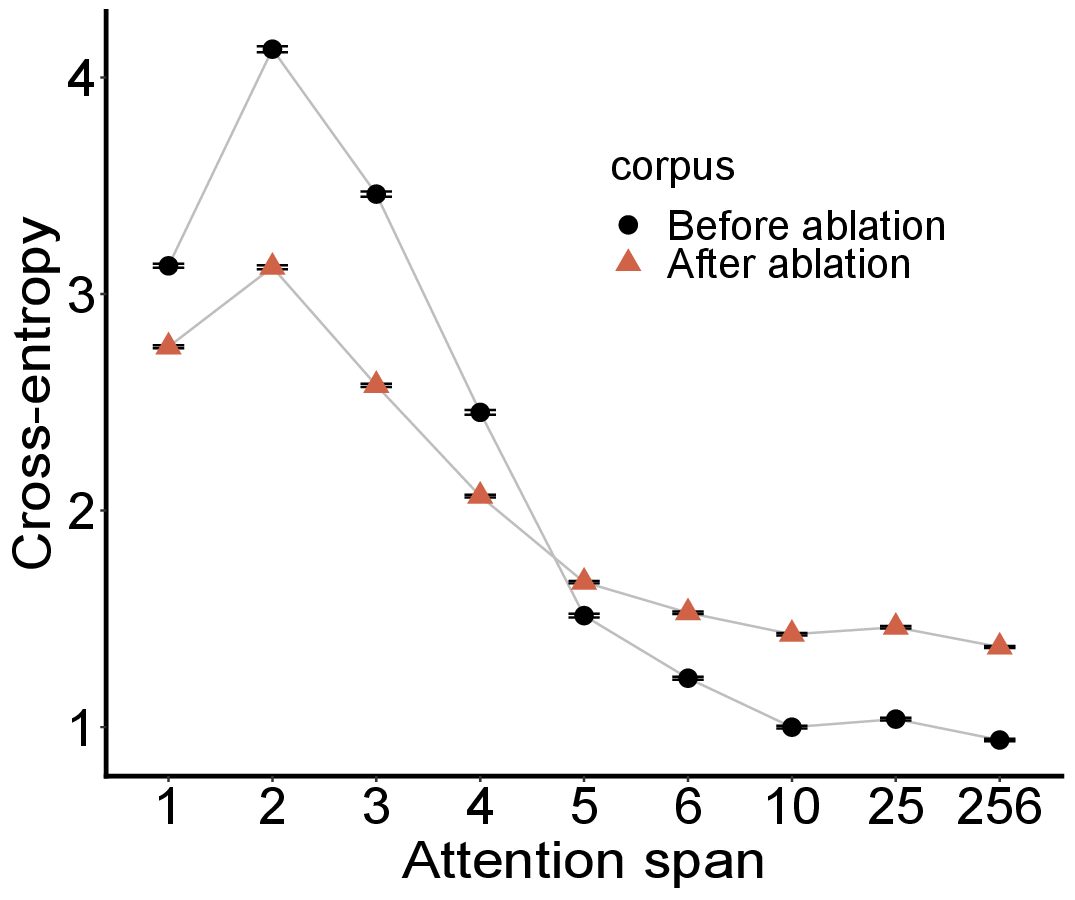}
			        \put(-7,80){\fontsize{10}{22}\selectfont\textbf{\textsf{E}}}
		   	   \end{overpic}	
				\hspace{3mm} % 横方向の空白を追加
			   \begin{overpic}[width=0.45\textwidth]{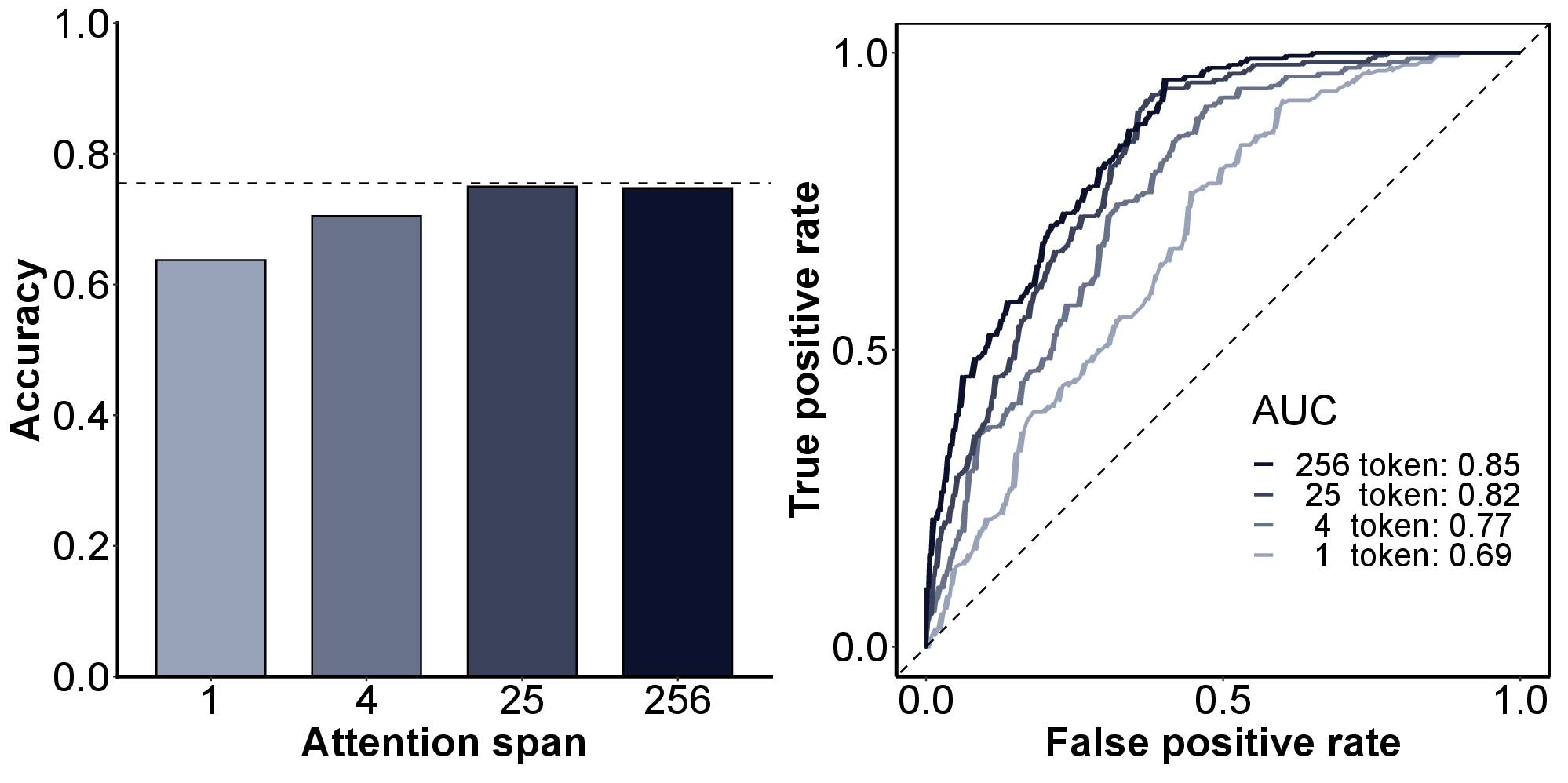}
			        \put(0,48){\fontsize{10}{22}\selectfont\textbf{\textsf{F}}}
		   	   \end{overpic}	
	  	\end{minipage}
  		    \centering

		    \caption{
\textbf{Ablation of brain nucleus reduces the performance of FinchGPT.}
\textbf{(A)} Schematic showing the song related nucleus of the songbird brain. Circuit diagrams show analogous function of songbird nucleus and corresponding human brain. Arrow indicates axonal projections. LMAN, lateral magnocellular nucleus of the aniterior nidopallium: DLM, dorsal lateral nucleus of the medial thalamus; nXIIts, XII cranical nerve; LMC, laryngeal motor cortex; LSC, laryngeal somatosensory cortex. 
\textbf{(B)} Example of the spectrograms of songs from identical birds before and after HVC ablation. 
\textbf{(C)} Example of spectrograms of syllables before and after HVC ablation (left) and the comparison of pitch values (right). Each point represents one syllable. Scale bar, 50 ms. P values, Wilcoxon signed-rank test, n = 13.
\textbf{(D)} Cross-entropy of FinchGPT-medium (6L, 6A, 384H) in the next-token prediction task, comparing the performance on before and after HVC ablation corpora. The model was trained on the before HVC ablation corpus, and the results were evaluated using the holdout method. Box plot shows median and first and third quantiles with mean shown as circles, P values, Wilcoxon rank sum test. n = 38,599 times prediction within 752 songs recorded before ablation, and n = 59,111 prediction across 1,180 songs recorded after ablation.
\textbf{(E)} The cross-entropy values of the model restricted to specific attention span lengths were analyzed for the next-token prediction task for songs from before and after HVC ablation (mean ± SEM, n = 38,599 and 59,111 predictions across 752 and 1,180 songs in the before- and after-ablation corpora, respectively).
\textbf{(F)} Comparison of performance FinchGPT with restricted Attention span length in classification of songs before and after HVC ablation. Accuracy (left), and ROC (Rate of change) and AUC (Area Under the Receiver Operating Characteristics Curve) (right). Dashed lines show the theoretical upper bound (left), and the chance level (right). The corpus of after HVC ablation was considered as positive in the ROC analysis. }
		    \label{fig:fig4}
		 \end{figure}

It have been suggested that the brain nuclei of songbirds have analogous regions in human brain with respect to their role in vocal control (Fig \ref{fig:fig4}A)~\cite{36, 37}. The pre-motor nucleus HVC (a letter-based name) play a pivotal role in song vocalization~\cite{38, 39}, playing an analogous role in speech production related areas including Broca’s area in human brain. It is known that HVC sends sequential and temporal information to the robust nucleus of the arcopallium (RA) which control muscle contraction in vocal organ for syllable articulation~\cite{40}. Disruption of normal function of neurons within HVC, such as through pharmacological ablation, is known to disrupt the sequential patterns of syllables in bird songs~\cite{39, 41, 42, 13}. Using this paradigm, we evaluated the model’s performance following HVC ablation.
 A song corpus was collected from the same bird both before and after hemispheric HVC ablation (Fig \ref{fig:fig4}B). HVC ablation disrupted normal ordering of songs without affecting the phonological features of syllables (Fig \ref{fig:fig4}C). FinchGPT, trained on before-ablation data, exhibited reduced performance when tested on after-ablation sequences in a next-token prediction task (Fig \ref{fig:fig4}D). Restricting the attention span from the full length of 256 resulted in an increase in cross-entropy, with a pronounced rise observed between 3–10 syllables in the before-ablation corpus, compared to the after-ablation corpus, which exhibit only a gradual increase (Fig \ref{fig:fig4}E). This suggests that long-range dependencies, particularly those involving relationships between 3–10 distal syllables, are disrupted by HVC ablation in this bird. 
We next conducted a song classification task in which FinchGPT, trained on before ablation songs, was tasked with distinguishing whether provided songs originated from corpora recorded before or after ablation of HVC. The accuracy of the song discrimination task decreased as the attention span was restricted (Fig \ref{fig:fig4}F). When the model was manipulated to reference only a single syllable, accuracy dropped significantly, indicating that that model could not effectively discriminate between before and after ablation songs. Restricting the attention span to 25 syllables, however, resulted in neglectable changes in accuracy compared to the full attention span of 256 syllables. AUC analysis further revealed a gradual decline in performance as the attention span was reduced (Fig \ref{fig:fig4}F). These findings suggest that songs from after ablation corpora exhibit a loss of long-range dependencies within syllables, in contrast to before-ablation songs.

Together, these results suggest that FinchGPT accurately learns the ordering of syllable sequence within natural birdsongs which was disrupted in HVC ablated birds.

\section{Discussion}
This study demonstrated the application of LLMs to birdsongs, offering a novel perspective on their structural properties. Traditionally, it has been postulated that language acquisition requires specific hardwired circuits that innately process hierarchical dependencies shared by all human languages~\cite{43}. Based on this assumption, the sequential rules in animal vocalization have remained enigmatic, as such rules are not shared between humans and non-human animals. However, recent advances in natural language processing have challenged this notion, demonstrating that language acquisition can occur through statistical learning without preassigned rules, although whether such models genuinely “understand” language in the same way as humans remains a subject of debate~\cite{44}.  This rule-free approach is particularly suited for identifying the statistical dependencies of vocal elements within animal vocalizations. By systematically comparing multiple algorithms, we identified Transformer-based models as exhibiting superior performance in modeling birdsongs. The Transformer  architecture, known for its ability to effectively leverage information across non-adjacent tokens and capture hierarchical structure rather than linear sequences~\cite{45}. The superior performance of Transformers suggests the presence of non-adjacent dependencies within birdsongs, a phenomenon we partially elucidated through attention layer analyses. 

The application of well-trained LLMs to animal vocalizations provides a robust framework to investigate the structural patterns of sequences in vocalizations. Beside the songbirds, higher-order rules within vocal elements have also been observed in some non-human animal species such as cetaceans~\cite{46, 47}, parrots~\cite{48}, elephants~\cite{49}, and non-human apes~\cite{50, 51}, although the details of them remain under debate. Future studies employing LLM-based analyses on vocal signals from these species could yield significant understanding of their underlying structure, further elucidating both the parallels and distinctions between human and animal communication systems. 
Being shown that LLM can capture the dependencies of elements within animal vocalization, an important future step is to compare how artificial neural network-mediated processing of animal vocalization aligns with actual neural processing in the brain of that animals. Cellular-level neural recording during natural recognition can be readily achieved in songbirds~\cite{40}, an experiment that is highly challenging not only in human brains but also in other large animals, such as whales and elephants. Such studies will contribute to understanding how “neurons” as computational units in artificial neural networks differ from and resemble the computations performed by actual neurons in the brains of living organisms~\cite{52, 53}. Conversely, further reverse engineering approach of FinchGPT, involving experiments that analyze and manipulate its attention mechanism, combining with analogous functional manipulation of neurons in the brain using cutting-edge technologies such as cell-specific optogenetic~\cite{54} and chemogenetic tools~\cite{55}, holds the potential to deepen our understanding of how language-related information is processed in animal brains. This research will shed light on the computational similarities and differences between artificial neural network models and biological neural networks in the processing of sequential data, especially those associated with language communications.

\section{Conclusion}
This study presents the development of a generative language model for birdsongs, demonstrating the superior performance of the Transformer algorithm compared to other models. Attention analysis visualized the computational process, revealing that the model effectively utilized long-range dependencies within the syllable sequence. Ablation of brain nucleus involved in song sequence generation led to diminished performance in models trained on pre-ablated song corpora. These findings underscore the potential of the Transformer architecture for analyzing and modeling animal communicative signals.

\section{Acknowledgments and funding sources}
We would like to thank Prof Ogawa for the discussion during the initial phase of this study. We thank the members of Abe Laboratory at Tohoku University for their help and fruitful suggestions. This research was supported by JSPS/MEXT KAKENHI(JP22H05482, JP24H01218, JP23K18252 JP21K19424) to KA, Tohoku University Research Program “Frontier Research in Duo” (Grant No. 2101) to KA and KI.

\section{Author contribution}
KA and KK conceptualized the project. KK performed all experiments and data analysis. KM, MT, KS and KI helped establish the language models. KA and KK wrote the initial manuscript. All the authors reviewed and edited the manuscript.

\end{document}